\documentclass[10pt,twocolumn,letterpaper]{article}

\usepackage{iccv}
\usepackage{times}
\usepackage{graphicx}
\usepackage{amsmath}
\usepackage{amssymb}

\usepackage{latexsym}
\usepackage{xspace}
\usepackage{caption}
\usepackage{subcaption} 
\usepackage{cprotect}
\usepackage{hhline}
\usepackage{array}
\usepackage{float}
\usepackage{xcolor}
\usepackage[group-separator={,},output-decimal-marker = {.}, group-minimum-digits={3}]{siunitx}  %
\usepackage{colortbl}
\usepackage{booktabs} %
\usepackage{multirow}
\usepackage[shortlabels]{enumitem} %
\usepackage[title]{appendix}

\usepackage[pagebackref=true,breaklinks=true,colorlinks,bookmarks=false]{hyperref}

\iccvfinalcopy %

\def\iccvPaperID{***} %

\ificcvfinal\fi

\usepackage{amssymb}
\usepackage{amsmath,amsfonts}
\usepackage{amsopn}
\usepackage{bm} %
\usepackage{multirow}
\newlength\savewidth

\newcommand{\ProbOpr}[1]{\mathbb{#1}}

\newcommand{\expect}[2]{%
\ifthenelse{\equal{#2}{}}{\ProbOpr{E}_{#1}}
{\ifthenelse{\equal{#1}{}}{\ProbOpr{E}\left[#2\right]}{\ProbOpr{E}_{#1}\left[#2\right]}}} %
\newcommand{\var}[2]{%
\ifthenelse{\equal{#2}{}}{\ProbOpr{VAR}_{#1}}
{\ifthenelse{\equal{#1}{}}{\ProbOpr{VAR}\left[#2\right]}{\ProbOpr{VAR}_{#1}\left[#2\right]}}} %

\newcommand{\eat}[1]{}

\newcommand{\mypartop}[1]{\vspace{0mm}\noindent\textbf{#1}.}
\newcommand{\mypar}[1]{\vspace{0.5em}\noindent\textbf{#1}.}

\newcommand{\coco}{COCO}

\newcommand{\oid}{Open~Images}
\newcommand{\oidcite}{\cite{kuznetsova20ijcv,openimages}}
\newcommand{\flickr}{Flickr30k}
\newcommand{\flickrcite}{\cite{young14tacl,plummer17ijcv}}
\newcommand{\ade}{ADE20K}
\newcommand{\cc}{Conceptual Captions}
\newcommand{\ccshort}{CC}
\newcommand{\cccite}{\cite{cc3m}}

\newcommand{\oidlocnar}{Open~Images Localized Narratives}
\newcommand{\oidlocnarshort}{OID LocNar}
\newcommand{\flickrlocnar}{\flickr{} Localized Narratives}
\newcommand{\flickrlocnarshort}{\flickr{} LocNar}

\newcommand{\adelocnarshort}{\ade{} LocNar}

\newcommand{\cocolocnarshort}{\coco{} LocNar}

\newif\ifdraft
\draftfalse
\ifdraft
  \newcommand{\beer}[1]{{\color{cyan}[Beer] #1}\xspace}
  \newcommand{\Jordi}[1]{{\color{olive}[Jordi] #1}\xspace}
  \newcommand{\vitto}[1]{{\color{orange}[Vitto] #1}\xspace}
  \newcommand{\radu}[1]{{\color{red}[Radu] #1}\xspace}

\else
  \newcommand{\beer}[1]{}
  \newcommand{\Jordi}[1]{}
  \newcommand{\vitto}[1]{}
  \newcommand{\radu}[1]{}

\fi

\begin{document}

\title{\vspace{-2mm}Telling the What while Pointing to the Where:\\Multimodal Queries for Image Retrieval\vspace{-2mm}}

\author{
Soravit Changpinyo, Jordi Pont-Tuset, Vittorio Ferrari, Radu Soricut\\
Google Research\\
  {\tt\small\{schangpi,jponttuset,vittoferrari,rsoricut\}@google.com}}

\maketitle
\pagestyle{plain}
\thispagestyle{plain}

\begin{abstract}
Most existing image retrieval systems use text queries as a way for the user to express \emph{what} they are looking for.
However, fine-grained image retrieval often requires the ability to also express \emph{where} in the image the content they are looking for is.
The text modality can only cumbersomely express such localization preferences, whereas pointing is a more natural fit.
In this paper, we propose an image retrieval setup with a new form of multimodal queries, where the user simultaneously uses both spoken natural language
(the \emph{what}) and mouse traces over an empty canvas (the \emph{where}) to express the characteristics of the desired target image.
We then describe simple modifications to an existing image retrieval model, enabling it to operate in this setup. 
Qualitative and quantitative experiments show that our model effectively takes this spatial guidance into account, and provides significantly more accurate retrieval results compared to text-only equivalent systems.
\end{abstract}

\vspace{-2mm}
\section{Introduction}
\label{sec:intro}

Gargantuan amounts of pictures are taken and shared every day, at an ever accelerating pace. Finding the picture that one has in mind should be easier and faster than painfully scrolling through hundreds of pictures in a digital-camera roll.
Building effective \emph{image retrieval} systems for finding specific images among large collections is, therefore, of paramount importance.
To speed the search up, image retrieval systems build an \emph{index} that represents a collection of images by automatically analyzing their content~\cite{zheng18pami,noh17iccv,gordo16eccv,song17iccv,liu17cvpr,sangkloy16tog,wang14cvpr,wang16cvpr,faghri17bmvc,lee18eccv,lu19nips,li20aaai,chen20eccv}.

\begin{figure}
\resizebox{\linewidth}{!}{%
\includegraphics{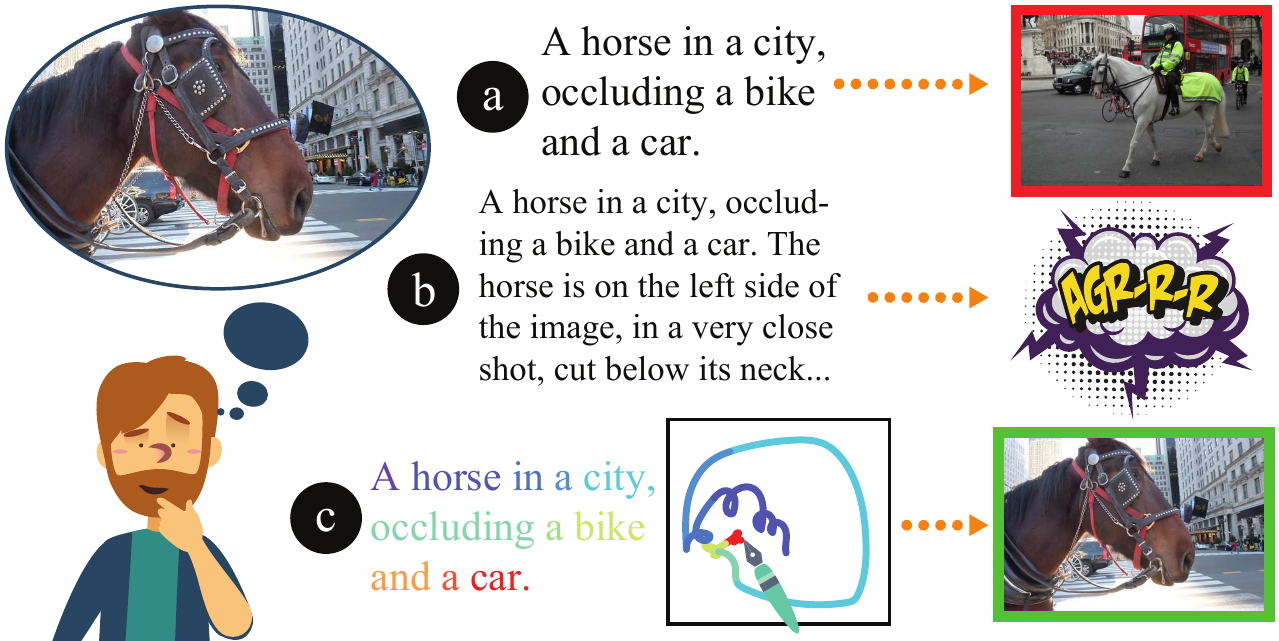}}
\caption{\small \textbf{Different types of textual queries} to represent the \emph{what} and the
\emph{where} in the target image: (a) spatial information is usually lacking in textual descriptions
and (b) it is cumbersome to express in written form, while (c) it is natural using mouse traces
synchronized with the text.}
\label{fig:splash}
\end{figure}

\begin{figure*}
  \centering
\resizebox{0.98\linewidth}{!}{%
\includegraphics{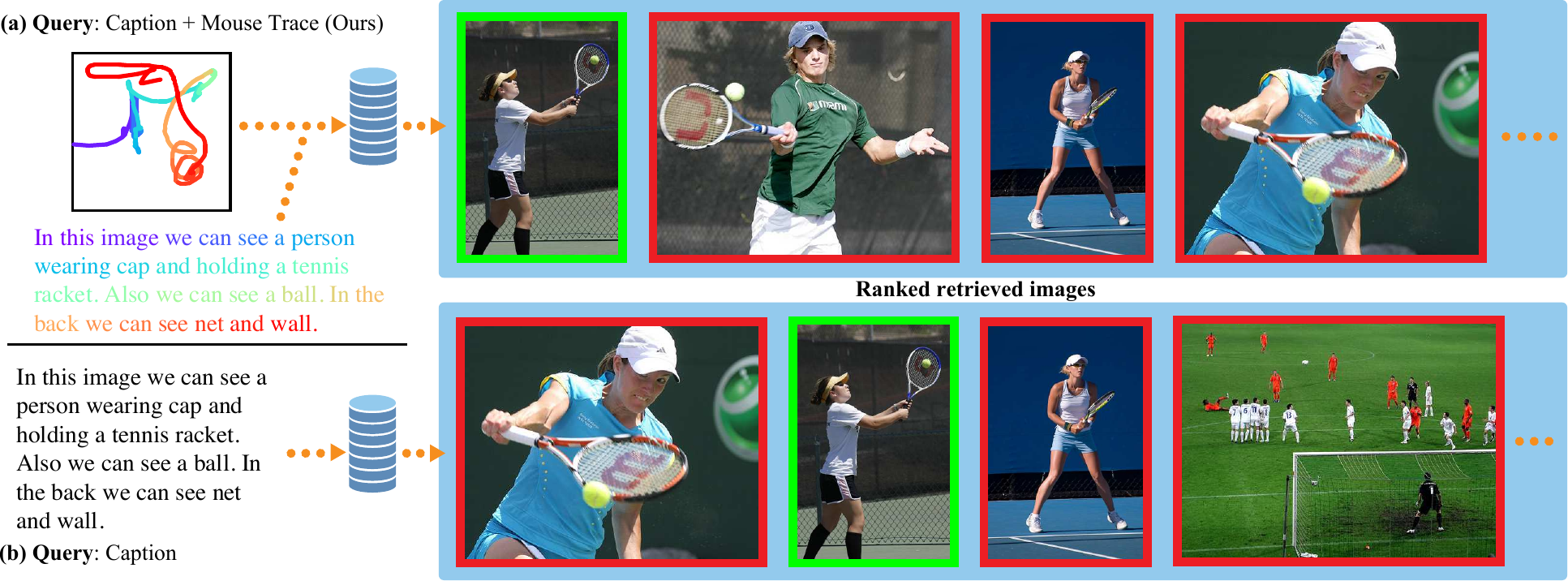}}
\vspace{-2mm}
\caption{\small\textbf{Qualitative results}: Querying with (a) text and mouse traces, versus (b) only text.
The target image is marked in green.
Adding mouse traces to express the spatial location of the image content allows us to get a better retrieval
result even given the same textual query.
In this particular case, notice that the exact position of the racket and the ball allow the model to
detect the correct target image.
\vspace{-2mm}}
\label{fig:intro_example}
\end{figure*}

A \emph{query} is a description of what a user is looking for in an image, a translation of their mental model of the target image
into a concrete form that can be understood by a retrieval system.
At a coarse level, a query can be a list of specific classes of objects (\eg{}, cars, people)
the user wants to be contained by the target image~\cite{torresani10eccv}.
At a finer-grained level is a natural language
description of its contents~\cite{wang14cvpr,wang16cvpr,faghri17bmvc,lee18eccv,lu19nips,li20aaai,chen20eccv}.
The latter is the most common paradigm in the recent literature, partly due to the availability of captioning datasets that can be used as training and testing data~\cite{lin14eccv,chen15arxiv,young14tacl,plummer17ijcv}.
These types of queries generally focus on \emph{what} is present in the image,
but fall short of expressing \emph{where} in the image the user expects it.

As an example, consider the image in Fig.~\ref{fig:splash}.
One textual query could be ``A horse in a city, occluding a bike and a car'' (Fig.~\ref{fig:splash}a).
The retrieved image, while not the one the user had in mind, is a perfect match for this description:
the \emph{what} in the image is similar to the intended target.
Expressing the \emph{where} part using the textual query is not only cumbersome for the user to write,
but also hard for the retrieval system to process (Fig.~\ref{fig:splash}b).

In this paper, we propose a new query modality where the user
describes the characteristics of the desired target image simultaneously using spoken
natural language, the \emph{what}, and mouse traces
over an empty canvas, the \emph{where} (Fig.~\ref{fig:splash}c).
Roughly pointing to an object's location comes naturally to humans~\cite{firestone14psyc,herbert05disc}
and is an effective way of communicating the image layout the user has in mind.
When the localization information is also temporally aligned with the natural language query,
it becomes a natural grounding signal that can be exploited to make retrieval more precise.

We propose an image retrieval model that takes this new type of multimodal query as input.
We start from an image-to-text matching model that is repurposed as an image retriever
by ranking image-text pairs according to their affinity, as in previous
literature~\cite{kiros2014unifying,faghri17bmvc,lee18eccv,zhang2020learning}.
We then augment the text input to also take the rough position in the blank canvas of each of the
words into account (Fig.~\ref{fig:model}).

The data for training and evaluating such a model comes from Localized Narratives~\cite{ponttuset20eccv}, a captioning dataset where 
annotators describe the images with their voice
while simultaneously moving their mouse over the objects
they are describing.
The mouse traces are effectively grounding each word of the caption in the image.
To use this data in an image retrieval scenario, we take the caption and corresponding mouse trace as input query, and the image on which the annotation is generated as target image.

Our experimental evaluation shows that this query modality provides a +$7$\% absolute better recall ($43$\% relative error rate decrease)
for the top image compared to a model using only text queries.
As we show in Fig.~\ref{fig:intro_example}, having the rough location of the objects mentioned in the query restricts the space of plausible images and thus allows for more effective retrieval results.

In summary, our main contributions are:
\begin{enumerate}[(a),nosep,leftmargin =* , widest* = 8]
  \item A novel query modality for fine-grained image retrieval that allows for a more natural specification of localization preferences.
  \item One concrete implementation of this idea that is simple and broadly applicable through a strong transformer-based model capable of incorporating the mouse traces.
  \item An experimental setup that suggests that Localized Narratives can be used to measure progress on this task.
  \item Empirical image retrieval results that demonstrate significant accuracy gains when the user is empowered with the ability to \emph{point to the where}.
\end{enumerate}

\section{Related Work}
\label{sec:related}

\begin{figure*}
\resizebox{1.0\linewidth}{!}{%
\includegraphics{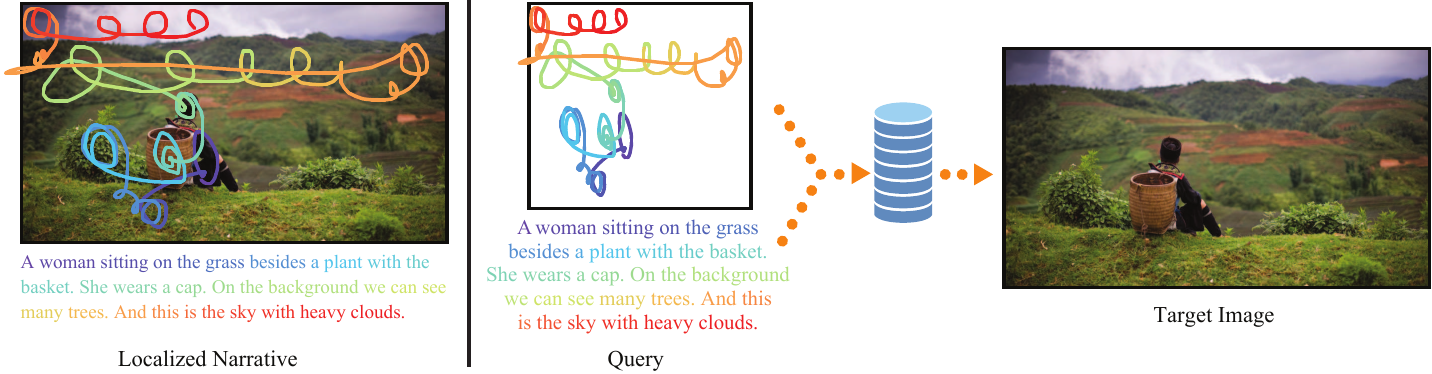}}
\caption{\small Localized Narratives annotations (left) can be transformed into training and testing data for image retrieval (right) by using the mouse traces as if they were drawn on a blank canvas, forming part of the query.}
\label{fig:locnarr}
\end{figure*}

\mypartop{Query Modality for Image Retrieval}
The closest line of work to ours is \textbf{text}-based image retrieval (discussed in detail below),
in which a natural language description serves as input to an image retrieval system.
We augment this input with mouse traces drawn on an empty canvas to express where in the image the content should appear.

Other works also augment the text query with a certain \textbf{structure} that indicates the \textit{where}, either limited to a closed vocabulary~\cite{johnson15cvpr,mai2017spatial,hinami2017region,furuta19mta,rossetto19arxiv,kilickaya21wacv}
or derived automatically from the natural language
descriptions~\cite{krishna17ijcv,li17cvprb,schuster15wvl,wang20wacv} (challenging in itself~\cite{xu17cvpr,li17iccvb,zellers18cvpr}).
In contrast, our mouse traces cover all words and are drawn as input.

Drawing \textbf{sketches} on an empty canvas~\cite{sangkloy16tog,song17iccv,liu17cvpr,bui18cg,zhang18eccvb} was also used to
represent an abstraction of an object category.
We argue that expressing the \emph{what} in natural language is significantly more intuitive and
faster than drawing a sketch (\eg{} compare using ``horse'' versus drawing one with enough detail
to differentiate it from a zebra).

In content-based image retrieval, the query is an \textbf{image} and the target image depicts either
(i) the same
object~\cite{zheng18pami,radenovic18pami,noh17iccv,gordo16eccv}, typically from another viewpoint,
at another time of the day, \etc (instance-level);
or
(ii) another object of the same category~\cite{chatfield14accv,sharma15iccv,newsam01icip} (category-level).
One can also add some natural language text that describes the desired modifications to the input
image~\cite{guo2018dialog,vo19cvpr,chen2020image}.
However, querying by image is a rather inflexible way to express what the user has in mind, as it already has its
content fixed (both the what and the where).

We believe that our query modality makes the most efficient use of both natural language and mouse traces:
the former to express a fine-grained \emph{what} naturally and fast, and the latter to specify the \emph{where} effectively and intuitively.

\mypar{Approach to Caption-Based Image Retrieval}
We focus on the most relevant works to ours, given the vast literature~\cite{zhou2017recent,chen2021deep} .
Typical methods learn deep representations of images and texts, fuse them, and score the fused representations.
To this end, a variety of factors have contributed to retrieval performance, including image and text features and
encoders, types of cross-modal interaction, approaches to hard negative mining, loss functions,
and pre-training data sources. \cite{burns2019language,chen2021learning,hendricks2021decoupling} investigate the effects of these factors.

Convolutional and recurrent neural networks with late fusion were popular in earlier works~\cite{wang14cvpr,kiros2014unifying,karpathy2015deep,klein2015associating,plummer2015flickr30k,ma2015multimodal,faghri17bmvc,huang17cvpr,zhang18eccv}, whereas recent works use transformers~\cite{lu19nips,li20aaai,chen20eccv,lu2021visualsparta,zhang2020learning,messina2020transformer,messina2020fine}, graph neural networks~\cite{li2019visual,wang2020cross,diao2021similarity}, or architectures with more complex cross-modal interactions~\cite{lee18eccv,wang19iccv,chen2020imram}.
The latter often leverage region-based ``bottom-up" visual features~\cite{anderson18cvpr,plummer17ijcv,lee18eccv}.
Moreover, multiple losses are explored, often requiring image-text triplets and hard negative mining~\cite{wang16cvpr,faghri17bmvc,eisenschtat2017linking,nam2017dual,zhang18eccv,wei2020universal,chen2020adaptive}.
Finally, pre-training image retrieval systems with large-scale image-text data sources has been shown to be extremely beneficial~\cite{gong2014improving,lu19nips,li2019visual,li20aaai,chen20eccv,qi2020imagebert,changpinyo2021cc12m,radford2021learning,jia2021scaling}.

Our base image-text matching model (Sec.~\ref{sec:approach}) follows most recent work~\cite{lu19nips,li20aaai,chen20eccv} that uses transformers~\cite{vaswani2017attention} with region-based
Faster R-CNN visual features~\cite{ren2015faster} trained on Visual Genome~\cite{krishna17ijcv}.
In addition, we explore the use of the Conceptual Captions~\cite{cc3m}, following~\cite{lu19nips}, and Localized Narratives~\cite{ponttuset20eccv} as additional pre-training data sources.
Finally, we adopt late image-text fusion~\cite{kiros2014unifying,faghri17bmvc} due to its simplicity, scalability, and effectiveness over early-fusion-based approaches in scenarios where large-scale pre-training data and contrastive learning with a large batch size are used~\cite{changpinyo2021cc12m,radford2021learning,jia2021scaling}.

Building on top of our strong caption-based image retrieval system, our approach to connecting text tokens and image regions via box representations (Sec.~\ref{sec:approach}, the orange boxes in Fig.~\ref{fig:model}) is largely inspired by position/location embeddings that are used extensively in recent work from both the computer vision and NLP communities~\cite{vaswani2017attention,devlin19bert,lu19nips,ponttuset20eccv}.

\section{New Query Modality}
\label{sec:task}

\mypar{Description}
We propose a new query modality for image retrieval in which the user provides a mouse trace on a blank canvas and a
natural language description that are synchronized with each other.
This allows the user to seamlessly specify
\emph{what} they want (through language) and \emph{where} they want it (through mouse traces, Fig.~\ref{fig:splash}).
We argue that pointing is a more natural means for taking into account the user's spatial preferences than existing approaches (Sec.~\ref{sec:related}).

\mypar{\emph{What+Where} Image Retrieval Setting}
As a second contribution, we construct the setting of \emph{what+where} image retrieval, and leverage the recent
Localized Narratives~\cite{ponttuset20eccv} dataset for this purpose.
It is a collection of image-caption pairs, where each caption word is grounded in the image by a mouse trace segment (Fig.~\ref{fig:locnarr} left).
They were obtained by annotators describing the images with their voice while simultaneously moving their mouse over the objects they were describing.

We transform the original Localized Narratives into useful annotations for image retrieval, by forming a query-image pair for each Localized Narrative as follows.
We first strip away the image and keep only the caption and synchronized mouse trace, as if it
had been drawn on an empty canvas.
This forms our input query.
Then we place the underlying image in our database, forming the intended target for that query (Fig.~\ref{fig:locnarr} right).

In the remainder of the paper we describe an image retrieval model that can operate in this setting (Sec.~\ref{sec:approach}), and then experimentally show that this leads to more accurate results with respect to the user's intent (Sec.~\ref{sec:exp}).

\section{Technical Model}
\label{sec:approach}

\begin{figure*}
\centering
\vspace{-12pt}
\resizebox{0.98\linewidth}{!}{%
\includegraphics{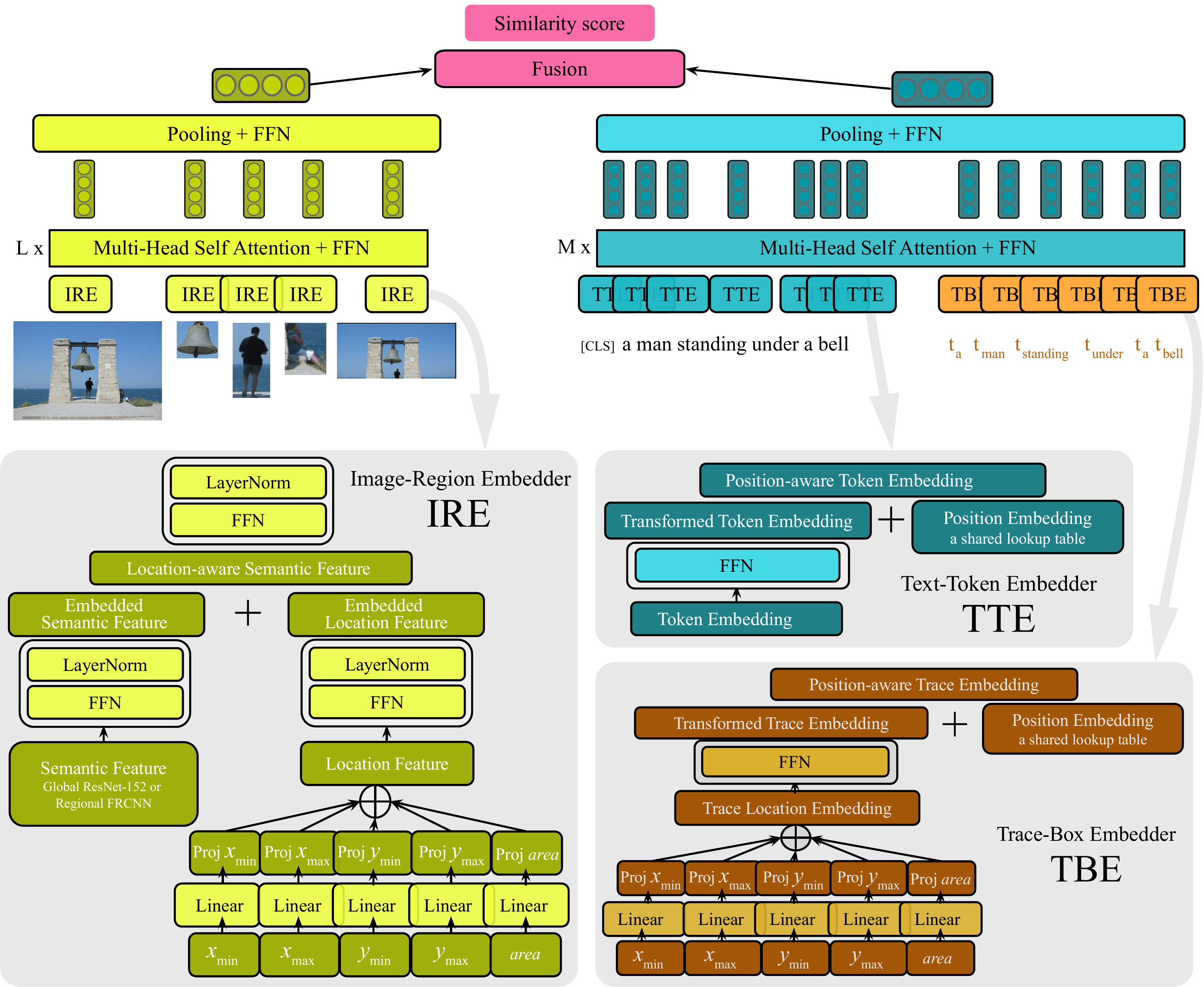}}
\vspace{-8pt}
\caption{\small\textbf{Model}: Our model performs early fusion of text token representations (blue) and the box representations
    (orange) using transformers.
Similarly, the model embeds the global and regional image embeddings (yellow).
During the late fusion, the model combines the two streams and computes the similarity score between the image embedding and
the text+traces embedding.}
\vspace{-10pt}
\label{fig:model}
\end{figure*}

In this section, we describe an approach that enables a strong image retrieval system to operate in the \emph{what+where} setting (Sec.~\ref{sec:task}).
We first describe our base image retrieval system based on image-text matching (Sec.~\ref{sec:basemodel}).
We then propose a modification to incorporate the extra input in the form of bounding boxes (Sec.~\ref{sec:basemodelwithtraces})
and show how we derive them from mouse trace segments (Sec.~\ref{subsec:trace2box}).

\subsection{Base image retrieval model}
\label{sec:basemodel}
As in much of the previous work (Sec.~\ref{sec:related}), we turn the standard text-based image retrieval problem into learning image-text matching.
Let us denote by $x = (x_1, \ldots, x_N)$ a set of feature vectors representing the image
(\eg{} the output of a CNN or an object detector run on the image) and
$y = (y_1, \ldots, y_K)$ a set of feature vectors representing the text 
(\eg{} random or pre-trained character/subword/wordpiece/word embeddings of text tokens).
We fix both $N$ and $K$ in our experiments and use padding and masking as necessary.

Our base model learns a similarity function
\begin{equation}
s(x, y) = p\big(f(x), g(y)\big),
\label{eq:imgtxtscore}
\end{equation}
where $f$, $g$, and $p$ are an image \textit{tower}, a text \textit{tower}, and an image-text \textit{fuser}, respectively.
Each tower reduces a set of feature vectors into a fixed-length vector and the fuser combines them to produce the final score.
In this paper, we choose the dot product as the image-text fuser $p$ and use the symmetric batched contrastive loss for parameter estimation, treating all other image-text pairs within the batch of size $B$ as negative examples:

\vspace{-0.2cm}
\begin{align}
L &= \frac{1}{2}(L_{x \xrightarrow{} y}+L_{y \xrightarrow{} x})\\
L_{a \xrightarrow{} b} &= \sum_i^B\log{\frac{\exp{(s(a^{(i)}, b^{(i)}))}}{\sum_{j=1}^{B} \exp{(s(a^{(i)}, b^{(j)}))}}}
\label{eq:loss} 
\end{align}
\vspace{-0.2cm}

At training time, we learn the parameters of $f$, $g$, and $p$ from a collection of image-text pairs.
At test time, given a query text $y'$,
we use the learned $p$
to compute a similarity score between $y'$
and each of the images $x$ in the database.
We then output a ranking of all database images sorted by their score, which represents our retrieval result.

Figure~\ref{fig:model} (without the trace inputs and the trace box embedder, in orange) illustrates our base model.
We adopt a two-stream model in which the image tower $f$ and the text tower $g$ do not share weights.
Each tower consists of three components: (i) an embedder, (ii) a contextualizer, and (iii) a pooler.
Both towers use a $6$-layer Transformer architecture~\cite{vaswani2017attention} for (ii) and mean pooling for (iii).
We use the vanilla architecture, where each transformer layer consists of a multi-head self-attention and feed-forward fully-connected network.
We refer the reader to \cite{vaswani2017attention} for details about the Transformer architecture.
Below, we describe the first component of each tower.

\mypar{The Image Region Embedder (IRE)} 
The input of the IRE is a fixed-length feature vector representing the whole image (CNN output) or a region of the image
(one of an object detector's region outputs).
The IRE transforms each of these feature vectors into an embedded semantic feature vector, and their corresponding 5D geometric feature 
of box coordinates ($x_{\min}$, $x_{\max}$, $y_{\min}$, $y_{\max}$) and box area into an embedded location feature.
Adding the two together gives a location-aware semantic feature vector of the region, which goes through a 2-layer Multi-Layer Perceptron (MLP)
before it is used as input of the image transformer.

\mypar{The Text Token Embedder (TTE)} 
Given a fixed-length vector representing a text token (a character, a subword, a word, \etc{}), the TTE applies a
2-layer MLP and adds a position embedding to the output, resulting in a token embedding that is position aware.  

We will use what we described here as our base image retrieval model throughout the paper, unless stated otherwise.
The end of Section~\ref{sec:related} discusses our modeling choices with respect to prior work.
Additionally, we verify that our implementation is strong, achieving a Recall@1 of \num{36.9} on the task of zero-shot image retrieval on \flickr{}~\flickrcite{} with \cc{}~\cite{cc3m} as pre-training data, outperforming ViLBERT~\cite{lu19nips}, a leading early-fusion, larger model.

\subsection{Incorporating mouse traces}
\label{sec:basemodelwithtraces}

Our high-level idea is to inject the traces to our base model by introducing the trace-box embedding (TBE) module whose encoded 1D text positions and 2D image locations act as a glue between text tokens and image regions.

Given mouse traces $t$ as an additional input, we modify our similarity function in~\eqref{eq:imgtxtscore} by injecting it into
the text stream of the model:
\begin{equation}
s(x, y) = p\big(f(x), h(y, t)\big),
\end{equation}
where $h$ is a text-trace fuser/embedder, and $f$ and $p$ are the same as in~\eqref{eq:imgtxtscore}.

Similarly to the setting in Section~\ref{sec:basemodel}, at training time we learn the parameters of $f$, $h$, and $p$ from a collection of
positive image-text-trace triplets.
At test time, given a query text $y'$
and its corresponding query trace $t'$, we use the learned $p$
to compute a similarity score between $(y',t')$
and each of the images in the database and output a ranking of the images.
Note that our setting assumes the existence of traces both during training and testing, as we envisage these new ``text+trace'' queries to
be cast by users using an interface analog to the one used for Localized Narratives annotation~\cite{ponttuset20eccv}.

Figure~\ref{fig:model} depicts our full model, with the components described in Section~\ref{sec:basemodel} unchanged.
The extra component, the mouse trace input $t$, is encoded in the form of a sequence of boxes by the Trace Box Embedder (TBE, bottom right of Fig.~\ref{fig:model}),
described below, and then fuse it with the text query.

\mypar{The Trace Box Embedder (TBE)}
Analogous to the location input of IRE, each of the trace boxes is represented using a 5D vector consisting of coordinates and area
($x_{\min}$, $x_{\max}$, $y_{\min}$, $y_{\max}$, $\mathit{area}$).
Since these boxes correspond to parts of the text query, they also have the notion of 1D time-location ``position'' in the query.
Thus, we add a position embedding to the transformed trace embedding vector, resulting in a trace embedding vector that is both
location-aware (visually) and position-aware (textually).

\mypar{Fusing texts and traces}
We concatenate all the outputs of TTE (Sec.~\ref{sec:basemodel}) and TBE, and use the result as input to the text-trace transformer.
We believe this is both simple and powerful, as the transformer self-attention layers allow text tokens and trace boxes to attend to
each other freely.
Note that it is this early fusion of text and traces that is capable of modeling where in the image certain parts of the query are
expected to be relevant.

\subsection{From mouse traces to bounding boxes}
\label{subsec:trace2box}
A Localized Narrative annotation has each utterance in the caption associated with a mouse
trace segment, which \emph{grounds} the utterance on the image.
In other words, it defines the rough position in the image where the semantic content from the utterance (the \emph{what}) is located (the \emph{where}).

The mouse trace segment for a certain utterance corresponds to the sequence of image points the mouse
traversed during the time interval ($t_1$, $t_2$) while the annotator spoke the utterance.
We observe that the mouse traces \textit{around} the time when an utterance was spoken can
still refer to the same utterance, so we explore adding \emph{temporal padding} $t_p$ to better
define the trace segment.
That is, we consider the trace segment in the time interval ($t_1-t_p$, $t_2+t_p$).

\begin{figure}
    \resizebox{1.0\linewidth}{!}{%
        \includegraphics{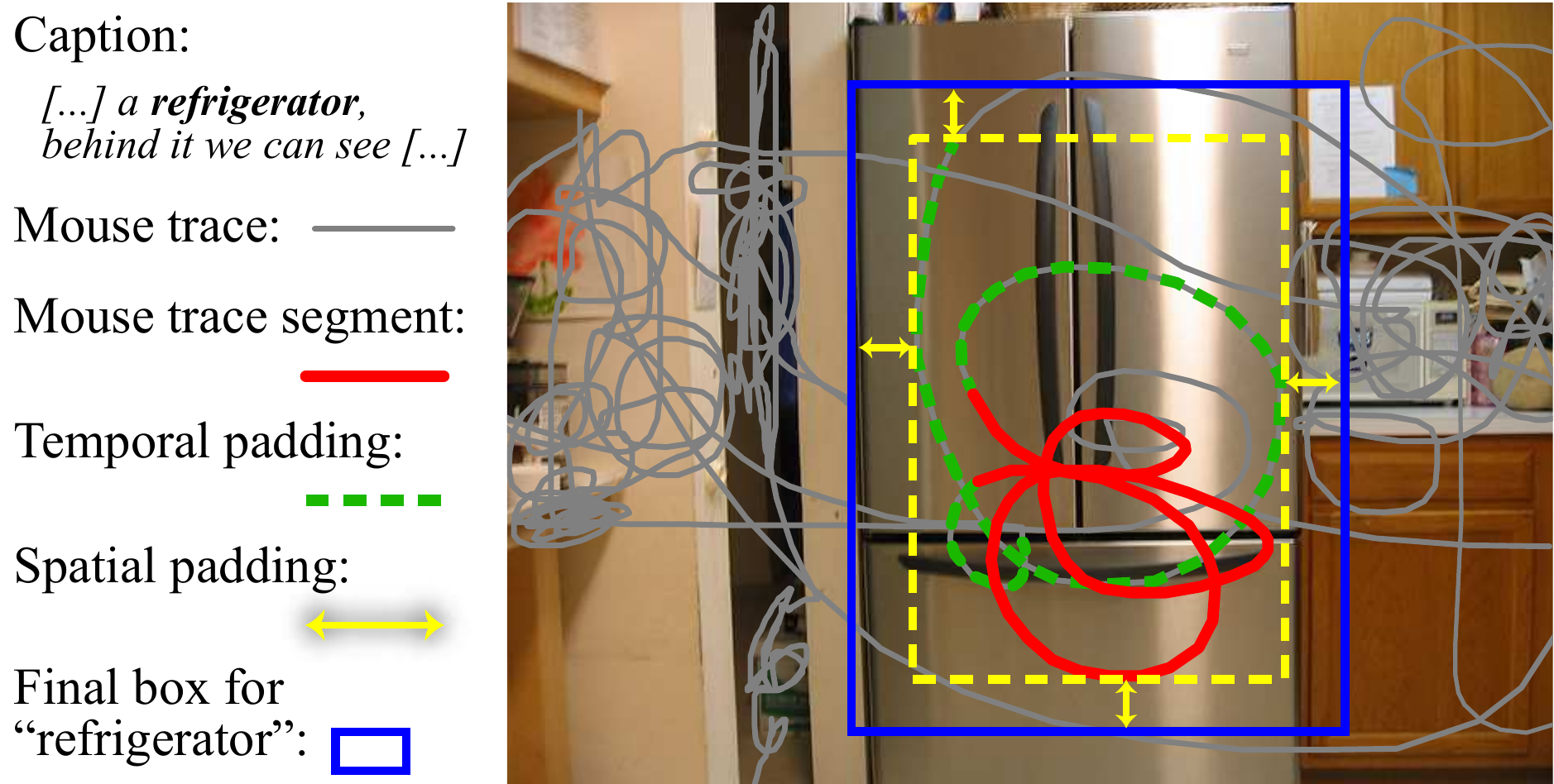}%
    }
    \caption{\small\textbf{From a mouse trace segment to its box}: We first prolong the mouse trace segment along the temporal dimension (green), and then we add spatial 2D padding (blue).}
    \label{fig:trace2box}
\end{figure}

As our model inputs bounding boxes that locate the query in the image (Fig~\ref{fig:model}),
we convert the mouse trace segments to boxes.
We start from the tightest box (Fig.~\ref{fig:trace2box}, yellow) that fully contains the
trace segment defined by the time segment ($t_1-t_p$, $t_2+t_p$), and we enlarge it in
all dimensions by a certain \emph{spatial padding} $s_p$ (Fig.~\ref{fig:trace2box}, blue).

\section{Experiments}
\label{sec:exp}


\subsection{Setup}
\label{sec:exp_setup}

\mypartop{Overview} The main goal of our experiments is to test whether incorporating mouse traces into the query improves the accuracy of image retrieval.
We will test this hypothesis in multiple scenarios, including several vision-and-language pre-training settings, inspired by~\cite{lu19nips,lu2012in1,li20aaai,chen20eccv}.

\mypar{Datasets}
Table~\ref{tab:data} summarizes the main datasets used in our experiments.
We use one dataset as our main task with multiple evaluation sets and two datasets as pre-training data sources. For the \textbf{main task}, we use \emph{\flickr{} Localized Narratives (\flickr{} LocNar)}~\cite{locnarr-website}.
This comes with the same set of \num{31783} images as \flickr{}, but we use the Localized Narratives captions and their synchronized mouse traces instead of the original captions without mouse traces.
We either train or fine-tune our models on the training split (\num{29783} images), perform model selection on the validation split (\num{1000} images), and report our quantitative results on the test split (\num{1000} images, Sec.~\ref{sec:exp_res_main}).
We further evaluate on different splits of \flickr{} and on two out-of-domain datasets: \emph{\coco{} Localized Narratives (\coco{} LocNar)} and \emph{\ade{} Localized Narratives (\ade{} LocNar)}~\cite{locnarr-website}, without additional fine-tuning (Sec.~\ref{sec:exp_res_detailed}).

For \textbf{pre-training}, we use the training splits of \emph{\cc{} (CC)}~\cccite{} and \emph{\oid{} Localized Narratives (\oidlocnarshort{})}~\cite{locnarr-website}.
The former contains \num{3.3}M pairs of (image, alt-text) harvested from the web.
The latter is a subset of the \num{9}M images in the \oid{} dataset~\oidcite{} that is annotated with Localized Narratives.
We use these annotations to pre-train both the image and the language branches of our model (Fig.~\ref{fig:model}).
We explore two pre-training data sources due to their complementary strengths:
\ccshort{} is larger-scale with more semantically specific terms (\eg{} croissant \vs food), while the style of descriptions in \oidlocnarshort{} is more similar to our target task of \flickrlocnarshort{}.
Furthermore, the existence of mouse traces in the \oidlocnarshort{} enables us to explore incorporating traces \emph{during pre-training} (using the model in Sec.~\ref{sec:basemodelwithtraces}).

\begin{table}
\small
\begin{center}
\begin{tabular}{c|c|c|c|}
Stage & Dataset & Size & \#Tok/cap \\ \hline
Main & \flickr{} LocNar & \num{31783} & \num{57.1} \\ \hline
Pretrain & \cc{} (train) & \num{3.3}M & \num{10.3} \\
Pretrain & \oid{} LocNar (train) & \num{507}K &  \num{35.5}\\ \hline
\end{tabular}
\vspace{-6pt}
\caption{\small \textbf{Main datasets} used in our experiments. LocNar is short for Localized Narratives. \#tok/cap is the average number of tokens per caption.}
\vspace{-15pt}
\label{tab:data}
\end{center}
\end{table}

\mypar{Settings}
We consider the from-scratch (no pre-training) setting and multiple pre-training settings: 
(i) on \ccshort{} only, (ii) on \oidlocnarshort{} only (with and without mouse traces), and (iii) on \ccshort{} followed by \oidlocnarshort{} (with and without mouse traces).
Setting (iii) is based on our intuition (which will be verified in the experiments) that the domain of \oidlocnarshort{} is closer to that of \flickrlocnarshort{}.

In each of these settings, we then compare the retrieval performance of the model with text-only queries (Sec.~\ref{sec:basemodel}) against that of the model with text+trace queries (Sec.~\ref{sec:basemodelwithtraces}) on the \flickrlocnarshort{}.
Note that when pre-training is involved, we make use of all available pre-trained weights and randomly initialize the rest (\eg{} the TBE weights when the mouse traces are not used during pre-training).

\mypar{Evaluation metrics} 
We use Recall@K (denoted as R@K for K=1,5,10):
the percentage of images in the test set for which the target image falls within the top-K of the model's output ranking, when using its corresponding text(+trace) as the input query.
We also report mean Average Precision (mAP) in our main experiments. Since we observe a consistent trend with that from R@K, we focus on R@K on the other experiments.

\mypar{Implementation details}
We use subtokens and random embeddings to represent text units (\eg{} ``standing'' $\xrightarrow[]{}$ ``stand'', ``ing'')
.
We use a vocabulary size of \num{10000}.
We represent an image with two types of features: A $2048$D global feature vector of ResNet152~\cite{resnet} and top $16$ regional feature vectors by a Faster-RCNN~\cite{ren2015faster} trained on Visual Genome~\cite{krishna17ijcv} with a ResNet101 backbone~\cite{resnet}.
Our box coordinates and area of a region are represented with relative numbers between 0 and 1, such that the 5D location information
$x_{\min}$, $x_{\max}$, $y_{\min}$, $y_{\max}$, and $\mathit{area}$ of the whole image is $0.0$, $0.0$, $1.0$, $1.0$, $1.0$, respectively.
We concatenate the two sets of features and permute the 16 regional vectors during training.
We use Adam~\cite{adam} and contrastive learning treating all other image-text pairs in each batch as negatives (Sec.~\ref{sec:basemodel}).
We tune an initial learning rate but always use a linear warm-up of $20$ epochs and multiply the learning rate by $0.95$ every $25$ epochs after that.

\subsection{Main Results}
\label{sec:exp_res_main}

\begin{table}[t!]
\small
\begin{center}
\begin{tabular}{c|c|ccc|c|}
\multicolumn{2}{c|}{Scenario} & \multicolumn{3}{c|}{Recall@K=} & mAP \\ \cline{1-2}
Pre-train? & Query & 1 & 5 & 10 & \\ \hline
& text & 63.5 & 87.4 & 92.8 & 74.0 \\
& text+trace & 68.2 & 88.8 & 94.4 & 77.7 \\ \hline
\checkmark & text & 83.4 & 97.6 & 98.5 & 89.7 \\
\checkmark & text+trace & \textbf{90.6} & \textbf{98.2} & \textbf{99.4} & \textbf{94.0} \\\hline
\end{tabular}
\vspace{-6pt}
\caption{\small \textbf{Main results.} The image retrieval performance on the \flickrlocnarshort{} 1K test set.}
\vspace{-15pt}
\label{tab:l2v_main}
\end{center}
\end{table}

\begin{table}[t]
\small
\begin{center}
\begin{tabular}{c|c|c|ccc|}
\multicolumn{2}{c|}{Pre-training} & Final & \multicolumn{3}{c|}{Recall@K=} \\ \cline{1-2}
data & query & query & 1 & 5 & 10 \\ \hline
\multirow{2}{*}{\ccshort{}} & text & text & 74.2 & 93.9 & 96.2 \\
& text & text+trace & 79.5 & 95.1 & 97.8  \\ \hline
 & text & text & 81.5 & 97.6 & 99.0  \\
{\footnotesize \oidlocnarshort{}} & text & text+trace & 83.9 & 97.1 & 98.5 \\
& text+trace & text+trace & \textbf{90.6} & 98.2 & \textbf{99.4} \\ \hline
\multirow{3}{*}{\shortstack{\ccshort{} $\xrightarrow{}$\\ {\footnotesize \oidlocnarshort{}}}} & text & text & 83.4 & 97.6 & 98.5 \\
 & text & text+trace & 83.5 & 97.2 & 98.2 \\
 & both  & text+trace & 90.2 & \textbf{98.4} & 99.0 \\ \hline
\end{tabular}
\vspace{-6pt}
\caption{\small \textbf{Pre-training with different data sources and query modalities} affects image retrieval performance on the \flickrlocnarshort{} 1K test set.}
\vspace{-15pt}
\label{tab:l2v_pretrain}
\end{center}
\end{table}

Table~\ref{tab:l2v_main} compares the image retrieval performance on \flickrlocnarshort{} of the models using the text-only queries and the ones using text+trace queries, i.e. our new \emph{what+where} setting (Sec.~\ref{sec:task}).

Regardless of whether we perform pre-training, incorporating the mouse trace (the ``where'') leads to significant gains in absolute R@1: $+4.7$\% without pre-training (Row~1 \vs{} Row~2), and $+7.2$\% with pre-training (Row~3 \vs{} Row~4).
Overall, the best result is obtained when we both pre-train and inject the trace to our model;
we improve over the baseline model by an absolute $+27.1$\%, $+10.8$\%, $+6.6$\% in R@\{1,5,10\}, and
$+20.0$\% in mAP (Row~1 \vs{} Row~4).

Our results suggest that \emph{the top retrieved image will be much more accurate if the user gets to ``point to the where''}.
Furthermore, \emph{pre-training and our new query modality are complementary}: the main benefit of pre-training with the text-only query modality is on improving ``telling the what''.

\subsection{Detailed Results and Ablation Studies}
\label{sec:exp_res_detailed}

\begin{figure*}
    \resizebox{1.0\linewidth}{!}{%
        \includegraphics{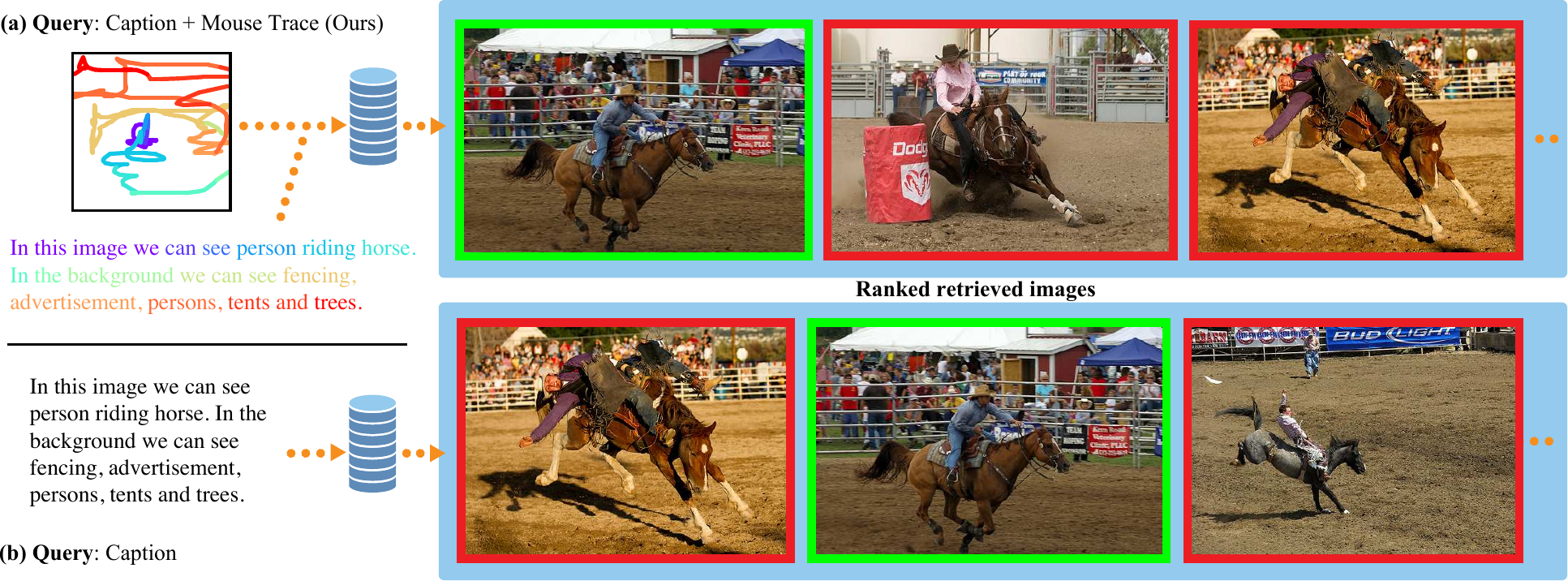}%
    }\vspace{-1mm}
    \caption{\small \textbf{Qualitative results}: Comparison between our best method (a) to that without trace supervision (b).
    In green, the target image that corresponds to the query on the left.}
    \label{fig:qualitative}
\end{figure*}

\mypartop{Pre-training data sources}
In Table~\ref{tab:l2v_pretrain}, we observe that \oidlocnarshort{} is superior to \ccshort{} as a pre-training data source for this task, supporting our intuition that the domain of the \oidlocnarshort{} is closer
to that of \flickrlocnarshort{}.
However, they are complementary when the trace is not involved.

\begin{table}[t!]
\small
\begin{center}
\begin{tabular}{c|c|ccc|}
Pre-training & Query & \multicolumn{3}{c|}{Recall@K=} \\
data & & 1 & 5 & 10 \\ \hline
\ccshort{} & text & 21.0 & 42.2 & 54.0 \\
\oidlocnarshort{} & text & 79.0 & 95.7 & 98.3 \\
\ccshort{} $\xrightarrow{}$ \oidlocnarshort{} & text & 79.1 & 95.7 & 97.9 \\ \hline
\oidlocnarshort{} & text+trace & 88.0 & 97.7 & 99.1 \\
\ccshort{} $\xrightarrow{}$ \oidlocnarshort{} & text+trace & 86.7 & 98.0 & 98.8 \\ \hline
\end{tabular}
\vspace{-6pt}
\caption{\small \textbf{Zero-shot image retrieval} performance on the \flickrlocnarshort{} 1K test set. Best viewed together with Table~\ref{tab:l2v_pretrain}.}
\vspace{-15pt}
\label{tab:l2v_zeroshot}
\end{center}
\end{table}

\begin{table}[t!]
\small
\begin{center}
\begin{tabular}{c|c|ccc|c|}
Eval data & Query & \multicolumn{3}{c|}{Recall@K=} & mAP \\
 & & 1 & 5 & 10 & \\ \hline
\multirow{2}{*}{\ade{}} & text & 47.4 & 73.8 & 84.6 & 59.5 \\
 & text+trace & \textbf{60.3} & \textbf{84.1} & \textbf{90.7} & \textbf{70.7}\\\hline
\multirow{2}{*}{\coco{}} & text & 73.7 & 94.3 & 97.6 & 82.5 \\
 & text+trace & \textbf{82.4} & \textbf{96.6} & \textbf{98.4} & \textbf{88.7} \\\hline
\end{tabular}
\vspace{-6pt}
\caption{\small \textbf{Out-of-domain evaluation.} Image retrieval performance on \adelocnarshort{} val (2K images) and \cocolocnarshort{} val (averaged over 5-fold 1K images).}
\vspace{-15pt}
\label{tab:l2v_more_datasets}
\end{center}
\end{table}

\mypar{Pre-training query modality}
In Table~\ref{tab:l2v_pretrain}, the largest benefit of pre-training is observed when text+traces are used during
both the pre-training and final stages;
in the case of \oidlocnarshort{}, this leads to the best R@1 of $90.6$.
When this is not possible (\ie{} pre-training data does not come with traces as in \ccshort{}), we still observe significant improvements in R@1 when using text+trace in the final stage.
This suggests that text+trace queries are \emph{generally} superior, working robustly across pre-training scenarios.

\mypar{Zero-shot image retrieval}
We test our models when they have not seen any image of the test domain (\flickrlocnarshort{}),
i.e. only trained on the pre-training data and evaluated on the \flickrlocnarshort{} test set (Tab.~\ref{tab:l2v_zeroshot}).
Together with Table~\ref{tab:l2v_pretrain}, we see that fine-tuning on \flickrlocnarshort{} is beneficial in all cases.
Notably, the zero-shot performance of the \ccshort{} model is much lower than the one fine-tuned on OID LocNar, indicating a big domain gap between \cc{} and Localized Narratives-style datasets.

\mypar{Out-of-domain evaluation}
We take the best text-only and text+trace models (last two rows of Tab.~\ref{tab:l2v_main}) as-is and evaluate their
performance on two additional datasets, \adelocnarshort{} and \cocolocnarshort{} (without fine-tuning on their training sets, Tab.~\ref{tab:l2v_more_datasets}).
The text+trace modality is still far superior to the text-only one ($+12.9$\% on \ade{} and $+8.7$\% in R@1 on \coco{}).
We stress that these datasets are in a different domain than the training sets (\oid{} and \flickr{}).
Thus, our improvements cannot be achieved simply by overfitting on the training domain.

\begin{table}[t]
\small
\begin{center}
\begin{tabular}{cc|cc|c|ccc|}
\multicolumn{2}{c|}{Image} & \multicolumn{2}{c|}{Text} & Trace & \multicolumn{3}{c|}{Recall@K=} \\ \cline{1-4}
sem & loc & tok & pos & & 1 & 5 & 10 \\ \hline
\checkmark & \checkmark & \checkmark & \checkmark & \checkmark & 68.2 & 88.8 & 94.4 \\ \hline
\checkmark & \checkmark & \checkmark & \checkmark & & 63.5 & 87.4 & 92.8 \\
\checkmark & \checkmark & & & \checkmark & 14.5 & 31.7 & 42.7 \\ \hline
\checkmark & & \checkmark & \checkmark & \checkmark & 66.8 & 89.4 & 94.5 \\
\checkmark & \checkmark & \checkmark & & \checkmark & 65.1 & 87.8 & 93.9 \\ \hline
\end{tabular}
\vspace{-6pt}
\caption{\small \textbf{Benefits of retrieval components} on the image retrieval performance on the \flickrlocnarshort{} 1K test set. The image features consist of semantic (sem) and 2D location (loc) embeddings. The text features consist of token (tok) and 1D position (pos) embeddings. See Sec.~\ref{sec:approach} and Fig.~\ref{fig:model} for details of these components.}
\vspace{-8pt}
\label{tab:l2v_ablation}
\end{center}
\end{table}

\mypar{Statistical significance}
We re-split the union of the training and test subsets of \flickrlocnarshort{} (keeping val intact) and then re-train and re-evaluate our best text+trace model (last row in Tab.~\ref{tab:l2v_main}).
Over 5 re-splits, the R@1 is $90.6\%\pm0.9$, which suggests that our gain of $+7.2$\% over the text-only model is statistically very significant.

\mypar{Trace-only query modality}
In Tab.~\ref{tab:l2v_ablation}, our trace-only query achieves a R@1 of \num{14.5} without pre-training (Row 3).
When compared to the text+trace and text-only queries (Row 1-2), this shows that while text plays a major role, both elements are important to achieve strong performance.

\mypar{Position and location embeddings}
Table~\ref{tab:l2v_ablation} also investigates the benefits of 1D word position (TTE) and 2D image region location (IRE) embeddings, both of which are connected to TBE in Figure~\ref{fig:model}.
We find that they are important as their absences lead to degradation in the top retrieved image (Row 1 \vs{} Row 4-5).

\mypar{Drawing traces on an empty canvas}
Analog to all modern text-to-image retrieval works that leverage image captioning datasets,
our experiments are limited by
the fact that our trace queries were drawn while the annotator was looking at the target image.
What if the traces were drawn on an empty canvas?
We select 7 images from the \flickrlocnarshort{} test set on which our best text+trace model retrieved the correct image in the top rank, but our best text-only model did not.
We then ask an annotator to briefly look at these 7 images, and then draw a trace for each image on an empty canvas, while reading the original caption ({\em without seeing the image}).
In this scenario, our text+trace model retrieves the correct image in 6 out of 7 cases, suggesting that our
model can maintain high accuracy even when the traces are not exactly aligned with the image regions.

\mypar{Architecture} In the supplementary material (Sec. B), we experiment with the number of layers of text (M) and image
(L) transformer encoders of our model (Fig.~\ref{fig:model}).
We find that the benefit of the text+trace query modality over the text-only one generalizes to all our ablation studies.

\mypar{Qualitative results} Figure~\ref{fig:qualitative} shows qualitative results, comparing our best model with text+trace query and our best model with text-only query.
Note that the exact positions of the fence and the advertisement allows the model to distinguish between images with very similar content.
More qualitative results are in Figure~\ref{fig:intro_example} and in the supplementary material (Sec. C).

\section{Conclusions}
\label{sec:discuss}

In this paper, we propose a new query modality for content-based image retrieval systems where the user
describes the characteristics of the desired target image simultaneously using spoken natural
language (the ``what'') and mouse traces over an empty canvas (the ``where'').
We present an image retrieval model that takes this new type of multimodal query as input.
We train and evaluate our model using Localized Narratives, where the caption and its corresponding mouse trace is used as input query, and the corresponding image as target.
Our experimental evaluation shows that this query modality provides a $~43$\% relative error rate decrease for the top image compared to the model that only uses text-based queries.

\section*{Acknowledgments}
We thank Sebastian Goodman for detailed feedback on an earlier version of the draft and code, Nan Ding, Piyush Sharma, and Ashish V. Thapliyal for help with coding, and Jason Baldridge for suggesting additional baselines. Some of the illustrations were downloaded from www.freepik.com (rawpixel.com, macrovector\_official).

{\small
\bibliography{arxiv_main}
\bibliographystyle{ieee_fullname}
}
\newpage
\begin{appendices}
\section{Additional Implementation Details}
\label{sec:app:details}

\subsection{Model}
We expand the details of our ``embedder," ``contextualizer," and ``pooler," described in the main text. Please refer to Figure~4 of the main paper for a high-level overview.

We use the same hyperparameter across Image-Region Embedder (IRE), Text-Token Embedder (TTE), and Trace-Box Embedder (TBE), described in the main text. Our feedforward network (FFN) is a $2$-layer MLP with the ReLU activation function, hidden size of $512$, and a dropout rate of $0.3$ (applied during training only). The dimension of position embedding (TTE, TBE) is set to $512$. To obtain a vector of size $512$ from a visual box or a trace box of $xmin, xmax, ymin, ymax, area$ (IRE, TBE). We perform a linear projection of this $5$D vector into $512$D and concatenate the result before feeding this $2560$D vector into FFN.

Our image (or text) transformer encoder has $6$ layers, a vocab embedding of size $512$, a hidden embedding of size $1024$, a filter size of $4096$. The number of attention heads is set to $8$.

Our image (or text) pooler is a mean pooling layer, followed by a $2$-layer MLP with the ReLU activation function, hidden size of $2048$, and a dropout rate of $0.3$ (applied during training only).  

\subsection{Learning}
We expand the description of our learning procedure in the main text.
For all experiments, we use Adam~\cite{adam} with default hyperparameters. We use Google Cloud \num{32}-core TPUs, with a batch size per core of \num{128}. The FRCNN regional features are permuted during training. We use a linear warmup of $20$ epochs, and a decay rate of $0.95$ every $25$ epochs for all experiments.

For all pre-training experiments, we use an initial learning rate of $0.000032$ 
The maximum number of training steps is set to \num{1}M for pre-training on \cc{} and to \num{250}K for pre-training on \oidlocnar{}.

We tune an initial learning rate when training or fine-tuning on \flickrlocnar{}. 
For from-scratch experiments --- with or without the mouse trace --- we find that a slightly higher learning rate of $0.000096$ works best. The maximum number of training steps is set to \num{70}K.

For fine-tuning experiments, there are two cases.
When both (latest) pre-training and fine-tuning stages use the same inputs, i.e., with or without the mouse trace in both stages, we use an initial learning rate of $0.0000032$, with one exception --- we find that \ccshort{}--pre-trained model requires a higher learning rate of $0.000032$ when fine-tuned on much longer captions from \flickrlocnarshort{}. We set the number of training steps to \num{10}K (without the mouse trace) or \num{25}K (with the mouse trace).
When the mouse trace is only incorporated during the fine-tuning stage, we observe better performance with slightly higher initial learning rates of $0.000096$ for \oidlocnarshort{}- and \ccshort{}$\xrightarrow[]{}$\oidlocnarshort{}--pre-trained models, and an even higher $0.00032$ for \ccshort{}--pre-trained model. We set the number of training steps to \num{70}K, as in the from-scratch experiments.

\begin{figure*}
    \resizebox{1.0\linewidth}{!}{%
        \includegraphics{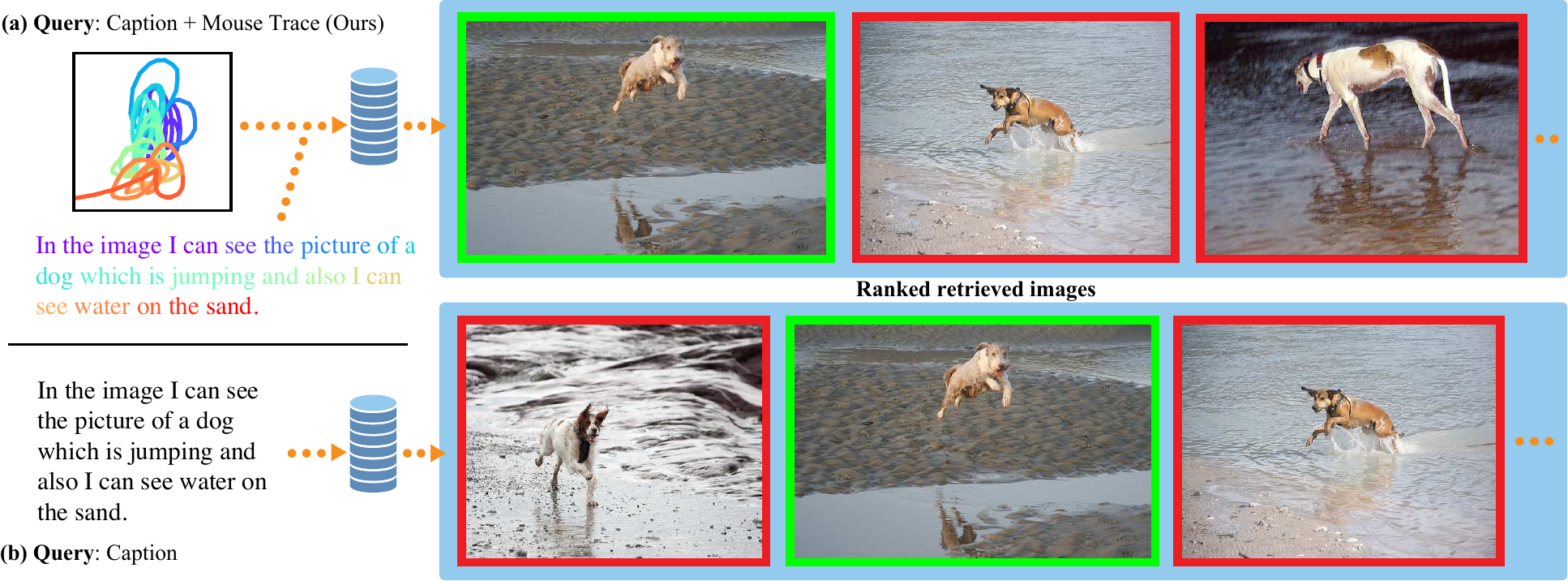}%
    }\vspace{1mm}
    \rule{\linewidth}{1pt}\\[4mm]
    \resizebox{1.0\linewidth}{!}{%
        \includegraphics{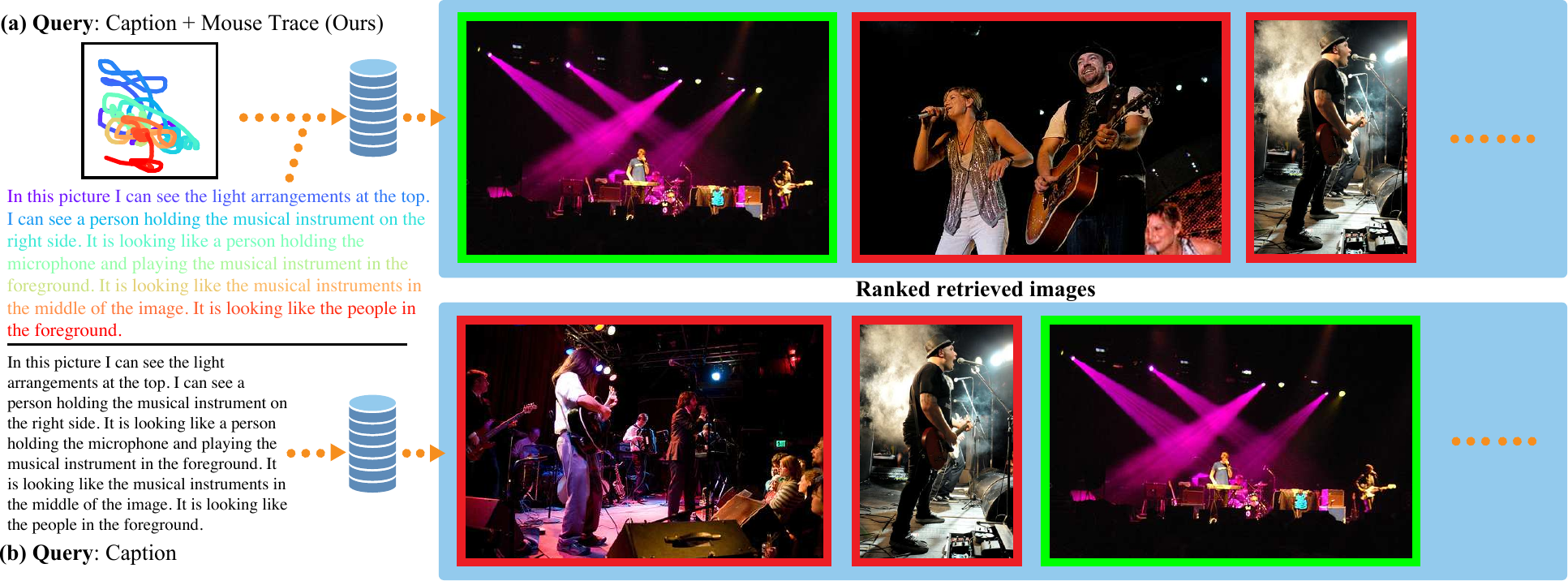}%
    }\vspace{1mm}
    \rule{\linewidth}{1pt}\\[4mm]
    \resizebox{1.0\linewidth}{!}{%
        \includegraphics{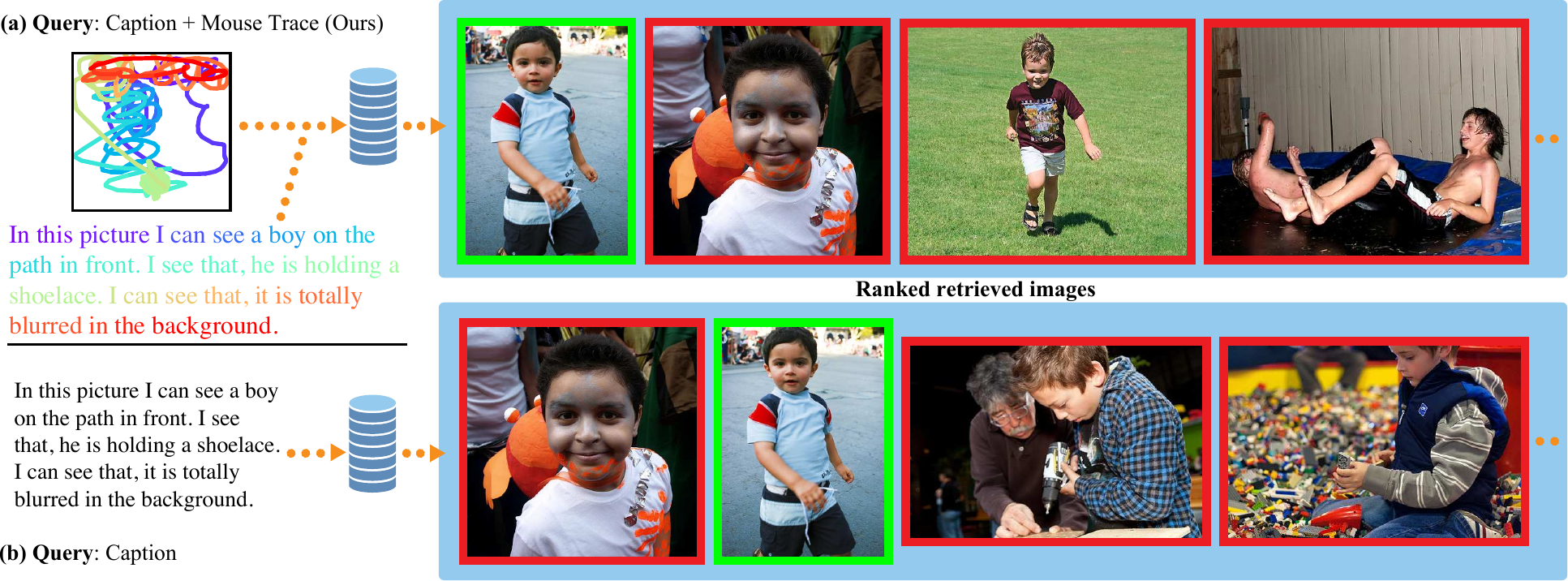}%
    }
    \caption{\small \textbf{Additional qualitative results (1 of 2)}:
    Comparison between the best model using our proposed text+trace query modality (a) to the best model using the text query modality (b).
    In green, the target image that corresponds to the query on the left.}
    \label{fig:qualitative1}
\end{figure*}

\begin{figure*}
    \resizebox{1.0\linewidth}{!}{%
        \includegraphics{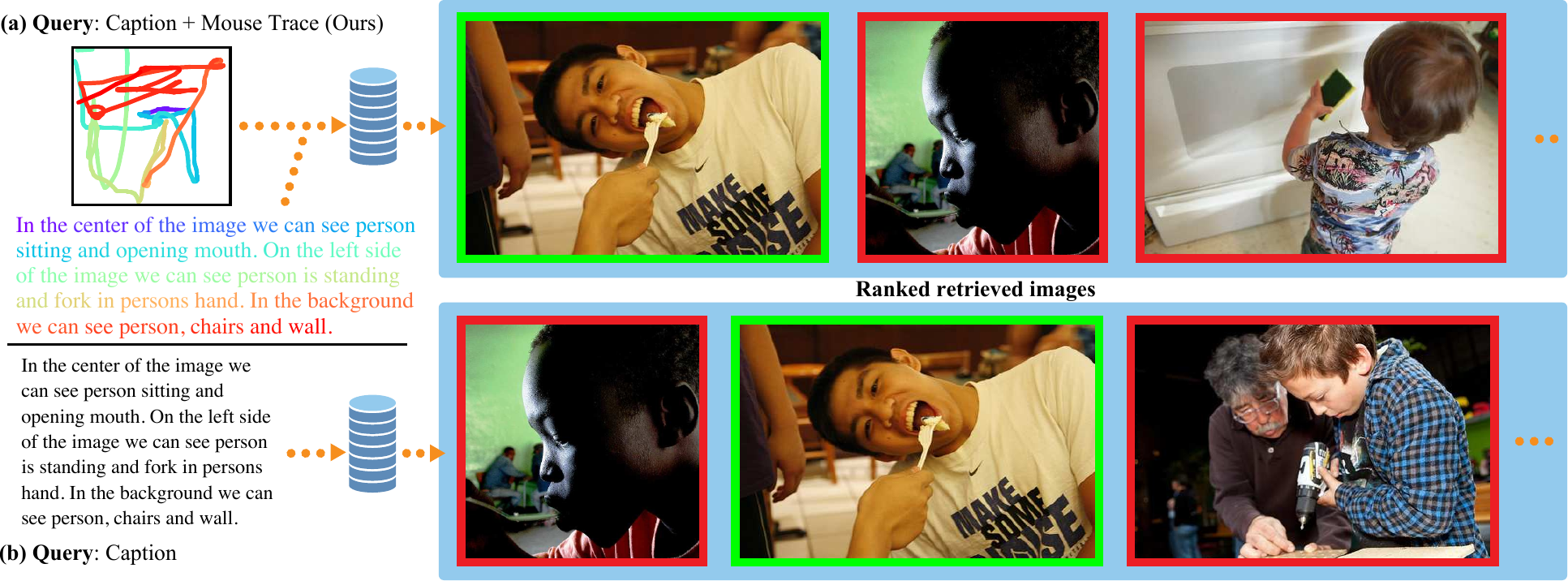}%
    }\vspace{1mm}
    \rule{\linewidth}{1pt}\\[4mm]
    \resizebox{1.0\linewidth}{!}{%
        \includegraphics{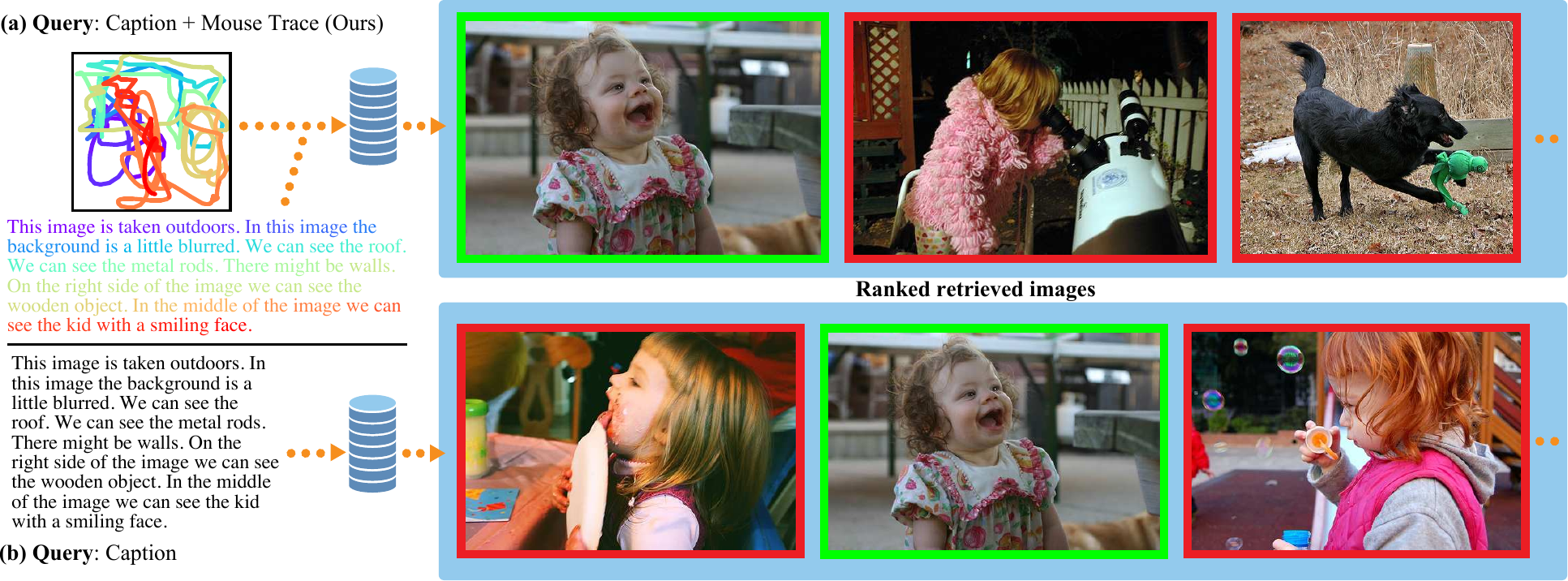}%
    }\vspace{1mm}
    \rule{\linewidth}{1pt}\\[4mm]
    \resizebox{1.0\linewidth}{!}{%
        \includegraphics{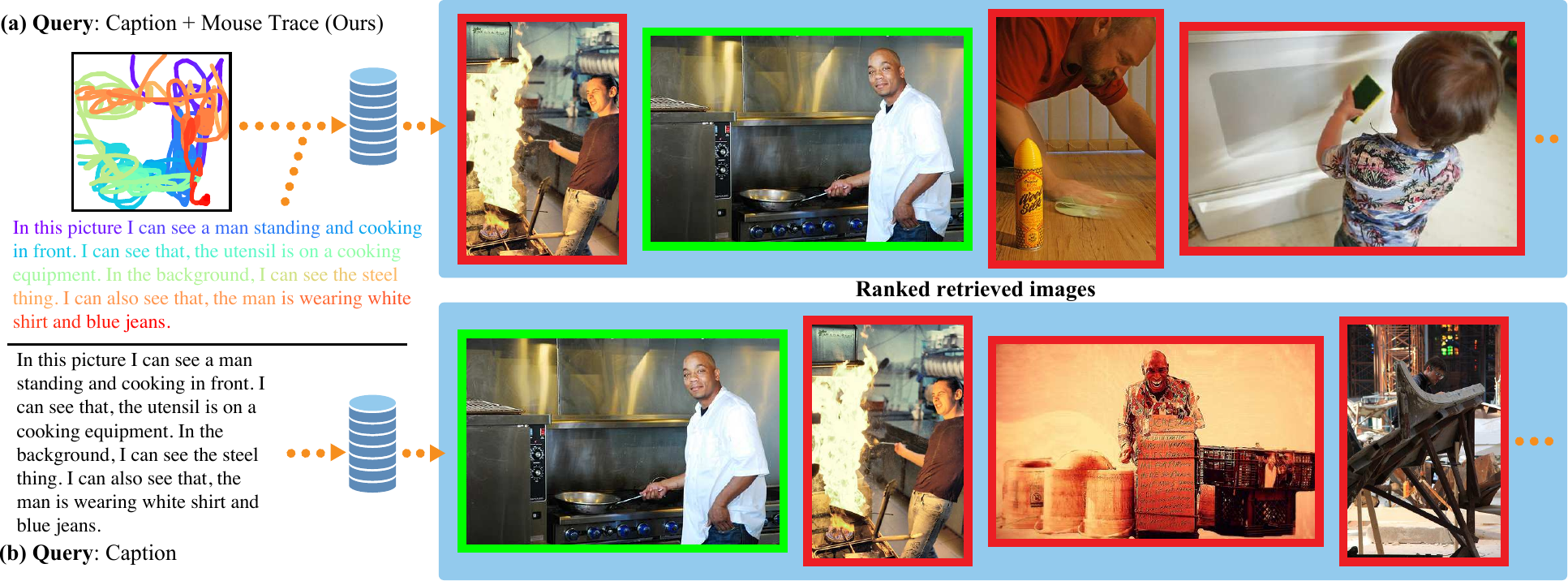}%
    }
    \caption{\small \textbf{Additional qualitative results (2 of 2)}:
    Comparison between the best model using our proposed text+trace query modality (a) to the best model using the text query modality (b).
    In green, the target image that corresponds to the query on the left.}
    \label{fig:qualitative2}
\end{figure*}

\section{Additional Experiments}
\mypar{Architecture} In Table~\ref{tab:l2v_architect}, we experiment with the number of layers of text (M) and image
(L) transformer encoders of our model (Fig.~4 in the paper).
We find that the benefit of the text+trace query modality over the text-only one generalizes to all our ablation studies.
Further, depth is more important in the text(+trace) branch than in the image one.

\begin{table}[t!]
\small
\begin{center}
\begin{tabular}{c|c|cc|c|cc|}
Query & Model & \multicolumn{2}{c|}{Recall@K=} & Model & \multicolumn{2}{c|}{Recall@K=} \\
 & L=6 & 1 & 5 & M=6 & 1 & 5 \\ \hline
text & M=6 & 63.5 & 87.4 & L=6 & 63.5 & 87.4 \\
text+trace & M=6 & 68.2 & 88.8 & L=6 & 68.2 & 88.8 \\ \hline
text & M=4 & 59.5 & 86.5 & L=4 & 61.7 & 87.2\\
text+trace & M=4 & 65.3 & 88.6 & L=4 & 64.8 & 87.9 \\\hline
text & M=2 & 58.3 & 84.0 & L=2 & 61.2 & 86.7 \\
text+trace & M=2 & 60.5 & 85.7 & L=2 & 64.6 & 88.6\\\hline
\end{tabular}
\vspace{-6pt}
\caption{\small \textbf{Benefits of depth:} the number of layers of text (M) and image (L) transformer encoders on the image retrieval performance on the \flickrlocnarshort{} 1K test set.}
\vspace{-8pt}
\label{tab:l2v_architect}
\end{center}
\end{table}

\section{Additional Qualitative Results}
\label{sec:app:results}

Figure~\ref{fig:qualitative1} and Figure~\ref{fig:qualitative2} show additional qualitative examples, comparing the best models from each modality.

As in the main text, the general trend is that using both the caption text and mouse trace (text+trace) is generally superior to the caption alone (text). For instance, text+trace correctly returns the target image with ``water on the sand" in Figure~\ref{fig:qualitative1} (top). We also find that the retrieval model struggles to use localization cues that are presented in the form of text such as ``at the top," ``on the right side," ``in the foreground," and ``in the middle of the image" in Figure~\ref{fig:qualitative1} (middle), representing the setting in Figure 1(b) of the main text. In many cases, the retrieval model fails to retrieve relevant images, for instance, ignoring ``shoelace" in Figure~\ref{fig:qualitative1} (bottom), ``person standing" in Figure~\ref{fig:qualitative2} (top), ``roof" in Figure~\ref{fig:qualitative2} (middle). However,  incorporating the trace helps correct this. We hypothesize that this is because the text+trace model is guided to focus on the right region, even in the presence of unfamiliar concepts and imperfect visual signals due to conditions such as poor lighting and extreme occlusion.

Finally, Figure~\ref{fig:qualitative2} (bottom) shows our failure case. We observe that one main mode of failure is that the text+trace model can be over-reliant to the trace and ignore part of the input text (``white shirt"); in this example, the trace of ``blue jeans," which covers a large space of the canvas, possibly misleads the model.

\end{appendices}

\end{document}